\documentclass[twocolumn,final]{elsarticle}

\usepackage{prletters}
\usepackage{framed,multirow}

\usepackage{amssymb}
\usepackage{latexsym}

\usepackage{url}	
\usepackage{xcolor}
\definecolor{newcolor}{rgb}{.8,.349,.1}

\usepackage{times}	
\usepackage{amssymb}
\usepackage{graphicx}
\usepackage{multirow}
\usepackage{pgfplots}
\usepackage{pdfpages}
\usepackage{float}
\usepackage{subcaption}
\usepackage{caption}

\usepackage[utf8]{inputenc}

\usepackage{blindtext}
\usepackage{color,soul}

\usepackage{verbatim}

\usepackage[acronym]{glossaries}
\usepackage{algorithm}
\usepackage{algorithmicx}
\usepackage{pdfpages}
\usepackage{time}
\usepackage{appendix}
\usepackage{amsmath}
\usepackage{amssymb}
\usepackage{sidecap}

\DeclareMathOperator{\Sim}{Sim}
\DeclareMathOperator{\adj}{adj}
\DeclareMathOperator{\ROUGEN}{ROUGE-N}
\DeclareMathOperator{\Count}{count}
\DeclareMathOperator{\match}{match}
\DeclareMathOperator{\gram}{gram}
\DeclareMathOperator{\ReferenceSummaries}{RS}

\newacronym{em}{EM}{Expectation Maximization}

\newacronym{pos}{POS}{Part-Of-Speech}
\newacronym{nlp}{PLN}{Processamento da Língua Natural}
\newacronym{svm}{SVM}{Support Vector Machine}
\newacronym{pca}{PCA}{Principal Component Analysis}
\newacronym{mfcc}{MFCC}{Mel-Frequency Cepstrum Coeficient}
\newacronym{bsr}{RSB}{Regra da Soma de Bayes}
\newacronym{me}{ME}{Máxima Entropia}
\newacronym{ner}{NER}{Named-Entity Recognition}

\newacronym{LSA}{LSA}{Latent Semantic Analysis}
\newacronym{TF-IDF}{TF-IDF}{Term Frequency - Inverse Document Frequency}
\newacronym{SVD}{SVD}{Singular Value Decomposition}

\newacronym{MMR}{MMR}{Maximal Marginal Relevance}
\newacronym{GRASSHOPPER}{GRASSHOPPER}{Graph Random-walk with Absorbing StateS
that HOPs among PEaks for Ranking}
\newacronym{ROUGE}{ROUGE}{Recall-Oriented Understudy for Gisting Evaluation}
\newacronym{KP-Centrality}{KP-Centrality}{Key Phrase-based Centrality}

\journal{Pattern Recognition Letters}

\begin{document}

\begin{frontmatter}

\title{Summarization of Films and Documentaries Based on Subtitles and Scripts}

\author[1,2]{Marta \snm{Aparício}}
\author[1,3]{Paulo \snm{Figueiredo}}
\author[1,3]{Francisco \snm{Raposo}}
\author[1,3]{David \snm{Martins de Matos}\corref{cor1}}
\ead{david.matos@inesc-id.pt}
\cortext[cor1]{Corresponding author}
\author[1,2]{Ricardo \snm{Ribeiro}}
\author[1]{Luís \snm{Marujo}}

\address[1]{L2F - INESC ID Lisboa, Rua Alves Redol, 9, 1000-029 Lisboa,
Portugal}
\address[2]{Instituto Universitário de Lisboa (ISCTE-IUL), Av. das Forças
Armadas, 1649-026 Lisboa, Portugal}
\address[3]{Instituto Superior Técnico, Universidade de Lisboa, Av. Rovisco
Pais, 1049-001 Lisboa, Portugal}

\received{00 May 2015}
\finalform{00 May 2015}
\accepted{00 May 2015}
\availableonline{00 May 2015}
\communicated{XXX}

\begin{abstract}

We assess the performance of generic text summarization algorithms applied to films and documentaries, using extracts from news articles produced by reference models of extractive summarization. We use three datasets: (i) news articles, (ii) film scripts and subtitles,  and (iii) documentary subtitles. Standard ROUGE metrics are used for comparing generated summaries against news abstracts, plot summaries, and synopses. We show that the best performing algorithms are LSA, for news articles and documentaries, and LexRank and Support Sets, for films. Despite the different nature of films and documentaries, their relative behavior is in accordance with that obtained for news articles. 

\end{abstract}


\end{frontmatter}


\section{Introduction}

Input media for automatic summarization has varied from text~\citep{luhn:1958,edmundson:1969} to 
speech~\citep{maskey:hirschberg:2005,zhang:chan:fung:2010,ribeiro2012_1427132508834}
and video~\citep{video:summ}, but the application domain has been, in general, restricted to informative sources:  news~\citep{barzilay:et:al:2002,radev:et:al:2005,conf/tsd/RibeiroM07,news:2014},
meetings~\citep{murray:renals:carletta:2005,garg:et:al:2009}, or
lectures~\citep{FujiiKN2007}.
Nevertheless, application areas within the entertainment industry are
gaining attention: e.g. summarization of literary short stories~\citep{KazantsevaS10}, music summarization~\citep{Raposo2015}, summarization of
books~\citep{mihalcea:explorations}, or inclusion of character analyses in
movie summaries~\citep{Sang2010}. We follow this direction, creating extractive, text-driven video summaries for films and documentaries. 

Documentaries started as cinematic portrayals of reality~\citep{DocsCloseReadings}. Today, they continue to portray historical events, argumentation, and research. They are commonly understood as capturing reality and therefore, seen as inherently non-fictional. Films, in contrast, are usually associated with fiction.   However, films and documentaries do not fundamentally differ: many of the strategies and narrative structures employed in films are also used in documentaries~\citep{realityIssuesAndConcepts}.   

In the context of our work, films (fictional) tell stories based on fictive events, whereas documentaries (non-fictional) address, mostly, scientific subjects. We study the parallelism between the information carried in subtitles and scripts of both films and documentaries. Extractive summarization methods have been extensively explored for news documents~\citep{LinHovy2000,McKeown2005,SparckJones2007,Radev01newsinessence,radev:et:al:2005,McKeown02trackingand}. Our main goal is to understand the quality of automatic summaries, produced for films and documentaries, using the well-known behavior of news articles as reference. Generated summaries are evaluated against manual abstracts using ROUGE metrics, which correlate with human judgements~\citep{Lin04rouge:a,LiuL2010}.

This article is organized as follows: Section~\ref{sec:generic-summarization}
 presents the summarization algorithms;
Section~\ref{sec:datasets} presents the collected datasets; Section~\ref{sec:setup} presents the evaluation setup; Section~\ref{sec:results} discusses our results; Section~\ref{sec:conclusions} presents conclusions and directions for future work.

\section{Generic Summarization}\label{sec:generic-summarization}	

Six text-based summarization approaches were used to summarize newspaper articles, subtitles, and scripts. They are described in the following sections.

\subsection{\gls{MMR}}\label{sub:mmr}

\gls{MMR} is a query-based summarization method~\citep{Carbonell1998}. It iteratively selects sentences via Equation~\ref{eq:e1} ($Q$ is a query; $\Sim_{1}$ and $\Sim_{2}$ are similarity metrics; $S_{i}$ and $S_{j}$ are non-selected and previously selected sentences, respectively). $\lambda$ balances relevance and novelty. \gls{MMR}
can generate generic summaries by considering the input
sentences centroid as a query~\citep{Murray2005,Xie2008}.
\begin{equation}\label{eq:e1}
\arg\max_{S_{i}}\left[\lambda{\Sim_{1}}\left(S_{i},Q\right)-\left(1-\lambda\right)\max_{S_{j}}\Sim_{2}\left(S_{i},S_{j}\right)\right]
\end{equation}

\subsection{LexRank}\label{sub:lexrank}

LexRank~\citep{Erkan2004} is a centrality-based method based on Google's
PageRank~\citep{Brin1998}. A graph is built using
sentences, represented by TF-IDF vectors, as vertexes. Edges are created when the cosine similarity exceeds a threshold. Equation~\ref{eq:lexrank} is computed at each vertex until the error rate between two successive iterations is lower than a certain value. In this equation, $d$ is a damping factor to ensure the method's convergence, $N$ is the number of vertexes, and $S\left(V_{i}\right)$ is the score of the $i$th vertex.
\begin{equation}\label{eq:lexrank}
S\left(V_{i}\right)=\frac{\left(1-d\right)}{N}+d\times\sum_{V_{j}\in
\adj\left[V_{i}\right]}\frac{\Sim\left(V_{i},V_{j}\right)}{\sum_{V_{k}\in
\adj\left[V_{j}\right]}\Sim\left(V_{j},V_{k}\right)}S\left(V_{j}\right)
\end{equation}

\subsection{\gls{LSA}}\label{sub:lsa}

\gls{LSA} infers contextual usage of text based on word co-occurrence~\citep{Landauer1997, landauer1998introduction}. Important topics are determined without the need for external lexical resources~\cite{Gong2001}: each word's occurrence context provides information concerning its meaning, producing relations between words and sentences that correlate with the way humans make associations.
\gls{SVD} is applied to each document, represented by a $t\times{n}$ term-by-sentences matrix $A$, resulting in its  decomposition $U\Sigma V^{T}$.
Summarization consists of choosing the $k$ highest singular values from $\Sigma$, giving $\Sigma_{k}$. $U$ and $V^{T}$ are  reduced to $U_{k}$ and $V^{T}_{k}$, respectively, approximating $A$ by $A_{k} = U_{k}\Sigma_{k} V^{T}_{k}$. The most important sentences are selected from $V^{T}_{k}$.


\subsection{Support Sets}\label{sub:support-sets}

Documents are typically composed by a mixture of subjects, involving a main and various minor themes. Support sets are defined based on this observation~\citep{Ribeiro2011}. Important content is determined by creating a support set for each passage, by comparing it with all others. The most semantically-related passages, determined via geometric proximity, are included in the support set. Summaries are composed by selecting the most relevant passages, i.e., the ones present in the largest number of support sets. 
For a segmented information source $I \triangleq p_1, p_2, \ldots, p_N$, support sets $S_i$ for each passage $p_i$ are defined by Equation~\ref{eq:e4}, where $\Sim$ is a similarity function, and $\epsilon_i$ is a threshold. The most important passages are selected by Equation~\ref{eq:e5}.
\begin{equation}\label{eq:e4}
 S_i \triangleq \{s \in I : \Sim(s,p_i) > \epsilon_i \wedge s \neq p_i\}
\end{equation}
\begin{equation}
\label{eq:e5}
\arg\max_{s \in U_{i=1}^n S_i} |\{S_i : s \in S_i\}|
\end{equation}

\subsection{\gls{KP-Centrality}}\label{sub:support-sets_luis}

\citet{ribeiro2013_1446676068775} proposed an extension of the centrality algorithm described in Section~\ref{sub:support-sets}, which uses a two-stage important passage retrieval method. The first stage consists of a feature-rich supervised key phrase extraction step, using the MAUI
toolkit with additional semantic features: the detection of rhetorical signals, the number of
Named Entities, \gls{pos} tags, and 4 n-gram domain model probabilities \citep{conf/interspeech/MarujoVN11,MARUJO12.672}. The second stage consists of the extraction of the most important passages, where key phrases are considered regular passages.

\subsection{\gls{GRASSHOPPER}}\label{sub:grasshopper}

\gls{GRASSHOPPER}~\citep{Zhu2007} is a re-ranking algorithm that maximizes diversity and minimizes redundancy. It takes a weighted graph $W$ ($n\times n$: $n$ vertexes representing sentences; weights are defined by a similarity measure), a probability distribution $r$ (representing a prior ranking), and $\lambda \in [0,1]$, that balances the relative importance of $W$ and $r$. If there is no prior ranking, a uniform distribution can be used.
Sentences are ranked by applying the teleporting random walks method in an absorbing Markov chain, 
based on the $n\times n$ transition matrix $\tilde{P}$ (calculated by
normalizing the rows of $W$), i.e., $ P=\lambda\tilde{P}+\left(1-\lambda\right)\textbf{1r}^{\top}$.
The first sentence to be scored is the one with the highest stationary probability $\arg\max_{i=1}^{n}\pi_{i}$ according to the stationary distribution of $P$: $\pi=P^\top\pi$.
Already selected sentences may never be visited again, by defining $P_{gg}=1$ and $P_{gi}=0,\forall i\neq g$. The expected number of visits is given by matrix $N=\left(I-Q\right)^{-1}$ (where $N_{ij}$ is the expected number of visits to the sentence $j$,
if the random walker began at sentence $i$).
We obtain the average of all possible starting sentences to get the
expected number of visits to the $j$th sentence, $v_{j}$. The sentence to be
selected is the one that satisfies $
\arg\max_{i=\left|G\right|+1}^{n}v_{i}$.

\section{Datasets}\label{sec:datasets}

We use three datasets: newspaper articles (baseline data), films, and documentaries. Film data consists of subtitles and scripts, containing scene descriptions and dialog. Documentary data consists  of subtitles containing mostly monologue. Reference data consists of manual abstracts (for newspaper articles), plot summaries (for films and documentaries), and synopses (for films).
Plot summaries are concise descriptions, sufficient for the reader to get a sense of what happens in the film or documentary. Synopses are much longer and may contain important details concerning the turn of events in the story. 
All datasets were normalized by removing punctuation inside sentences and timestamps from subtitles.

\subsection{Newspaper Articles}
TeMário~\citep{Pardo2003b} is composed by
100 newspaper articles in Brazilian Portuguese (Table~\ref{tab:news_corpus}), covering domains
such as ``world", ``politics", and ``foreign affairs". Each 
article has a human-made reference summary (abstract). 
\begin{table}[H]
	\caption{TeMário corpus properties.}
	\label{tab:news_corpus}
	\centering
	\begin{tabular}{llccc}
		\hline
		&           & AVG & MIN & MAX\\
		\hline
		\multirow{2}{*}{\#Sentences} & News Story & 29      & 12      & 68\\
		& Summary  & 9    & 5     & 18\\
		\hline
		\multirow{2}{*}{\#Words} &  News Story  & 608       & 421       & 1315\\

		& Summary  & 192     & 120       & 345\\
		\hline
	\end{tabular}
\end{table}

\subsection{Films}\label{sub:films}

We collected 100 films, with an average of 4 plot summaries (minimum of 1, maximum of 7) and 1 plot synopsis per film (Table~\ref{tab:filmsCorpus}).
Table~\ref{tab:filmsCorpus2} presents the properties of the subtitles, scripts, and the concatenation of both. Not all the information present in the scripts was used: dialogs were removed in order to make them more similar to plot summaries.

\begin{table}[H]
\caption{Properties of plot summaries and synopses.}
\label{tab:filmsCorpus}
\centering
\begin{tabular}{llccc}
\hline
                     &           & AVG & MIN & MAX\\
\hline
\multirow{2}{*}{\#Sentences} & Plot Summaries & 5       & 1       & 29\\
                     & Plot Synopses  & 89     & 6       & 399\\
\hline
\multirow{2}{*}{\#Words} & Plot Summaries & 107      & 14      & 600\\
                     & Plot Synopses  & 1677    & 221     & 7110\\
\hline
\end{tabular}
\end{table}

\begin{table}[H]
\caption{Properties of subtitles and scripts.}
\label{tab:filmsCorpus2}
\centering
\begin{tabular}{llccc}
\hline
                     &               & AVG & MIN & MAX\\
\hline
\multirow{3}{*}{\#Sentences} & Subtitles     & 1573    & 309     & 4065\\
                     & Script        & 1367    & 281     & 3720\\
                     & Script + Subtitles & 2787    & 1167    & 5388\\
\hline
\multirow{3}{*}{\#Words} & Subtitles     & 10460    & 1592    & 27800\\
                     & Script        & 14560   & 3493    & 34700\\
                     & Script + Subtitles & 24640   & 11690   & 47140\\
\hline
\end{tabular}
\end{table}

\subsection{Documentaries}\label{sub:documentaries}

We collected 98 documentaries. Table~\ref{table:Doc_Corpus} presents the properties of their subtitles: note that the number of sentences is smaller than in films, influencing ROUGE (recall-based) scores. 

\begin{table}[H]
\caption{Properties of documentaries subtitles.}
\label{table:Doc_Corpus}
\centering
\begin{tabular}{lccc}
\hline
            & AVG & MIN & MAX\\
\hline
\#Sentences & 340     & 212     & 656\\
\#Words     & 5864   & 3961   & 10490\\
\hline
\end{tabular}
\end{table}

We collected 223 manual plot summaries and divided them into four classes (Table~\ref{table:Doc_PlotSummaries}): 143  ``Informative", 63  ``Interrogative", 9  ``Inviting", and 8 ``Challenge". ``Informative" summaries contain factual information about the program; ``Interrogative" summaries contain questions  that arouse viewer curiosity, e.g. ``What is the meaning of life?"; ``Inviting" are invitations, e.g. ``Got time for a 24 year vacation?"; and, ``Challenge" entice viewers on a personal basis, e.g. ``are you ready for...?".
We chose ``Informative" summaries due to their resemblance to the sentences extracted by the summarization algorithms. On average, there are 2 informative plot summaries per documentary (minimum of 1, maximum of 3).

\begin{table}[H]
\caption{Properties of the documentary plot summaries.}
\label{table:Doc_PlotSummaries}
\centering
\begin{tabular}{llccc}
\hline
                     &               & AVG & MIN & MAX\\
\hline
\multirow{4}{*}{\#Sentences} & Informative   & 4       & 1       & 18\\
                     & Interrogative & 4       & 1       & 19\\
                     & Inviting      & 6       & 2       & 11\\
                     & Challenge     & 5       & 2       & 9\\
\hline
\multirow{4}{*}{\#Words} & Informative   & 82      & 26      & 384\\
                     & Interrogative & 103     & 40      & 377\\
                     & Inviting      & 146     & 63      & 234\\
                     & Challenge     & 104     & 59      & 192\\
                                          
\cline{1-5}
\end{tabular}
\end{table}

\section{Experimental Setup}\label{sec:setup}

For news articles, summaries were generated with the average size of the manual abstracts ($\approx31\%$ of their size).

For each film, two summaries were generated, by selecting a number of sentences equal to (i) the average length of its manual plot summaries, and (ii) the length of its synopsis. In contrast with news articles and documentaries, three types of input were considered: script, subtitles, script+subtitles.
	
For each documentary, a summary was generated with the same average number of sentences of its manual plot summaries ($\approx1\%$ of the documentary's size). 
	
Content quality of summaries is based on word overlap (as defined by ROUGE) between generated summaries and their references. ROUGE-N computes the fraction of selected words that are correctly identified by the summarization algorithms (cf. Equation~\ref{ROUGEN}: RS are reference summaries, $\gram_{n}$ is the n-gram length, and $\Count_{\match} (\gram_{n})$ is the maximum number of n-grams of a candidate summary that co-occur with a set of reference summaries). ROUGE-SU measures the overlap of skip-bigrams (any pair of words in their sentence order, with the addition of unigrams as counting unit). ROUGE-SU4 limits the maximum gap length of skip-bigrams to 4.
\begin{equation} \label{ROUGEN}
	\ROUGEN = { \sum_{S \in \ReferenceSummaries} \sum_{\gram_{n} \in S} \Count_{\match} (\gram_{n})
		\over \sum_{S \in \ReferenceSummaries} \sum_{\gram_{n}\in S} \Count(\gram_{n})}
\end{equation}

\section{Results and Discussion}\label{sec:results}


Subtitles and scripts were evaluated against manual plot summaries and synopses to define an optimal performance reference.
The following sections present averaged ROUGE-1, ROUGE-2, and ROUGE-SU4 scores (henceforth R-1, R-2, and R-SU4), and the performance of each summarization algorithm, as a ratio between the score of the generated summaries and this reference (relative performance). Several parametrizations of the algorithms were used (we present only the best results). Concerning MMR, we found that the best $\lambda$ corresponds to a higher average number of words per summary. Concerning GRASSHOPPER, we used the uniform distribution as prior. 

\subsection{Newspaper Articles (TeMário)}

\begin{figure}[H]
	\begin{center}
		\includegraphics[keepaspectratio=true,width=\columnwidth]{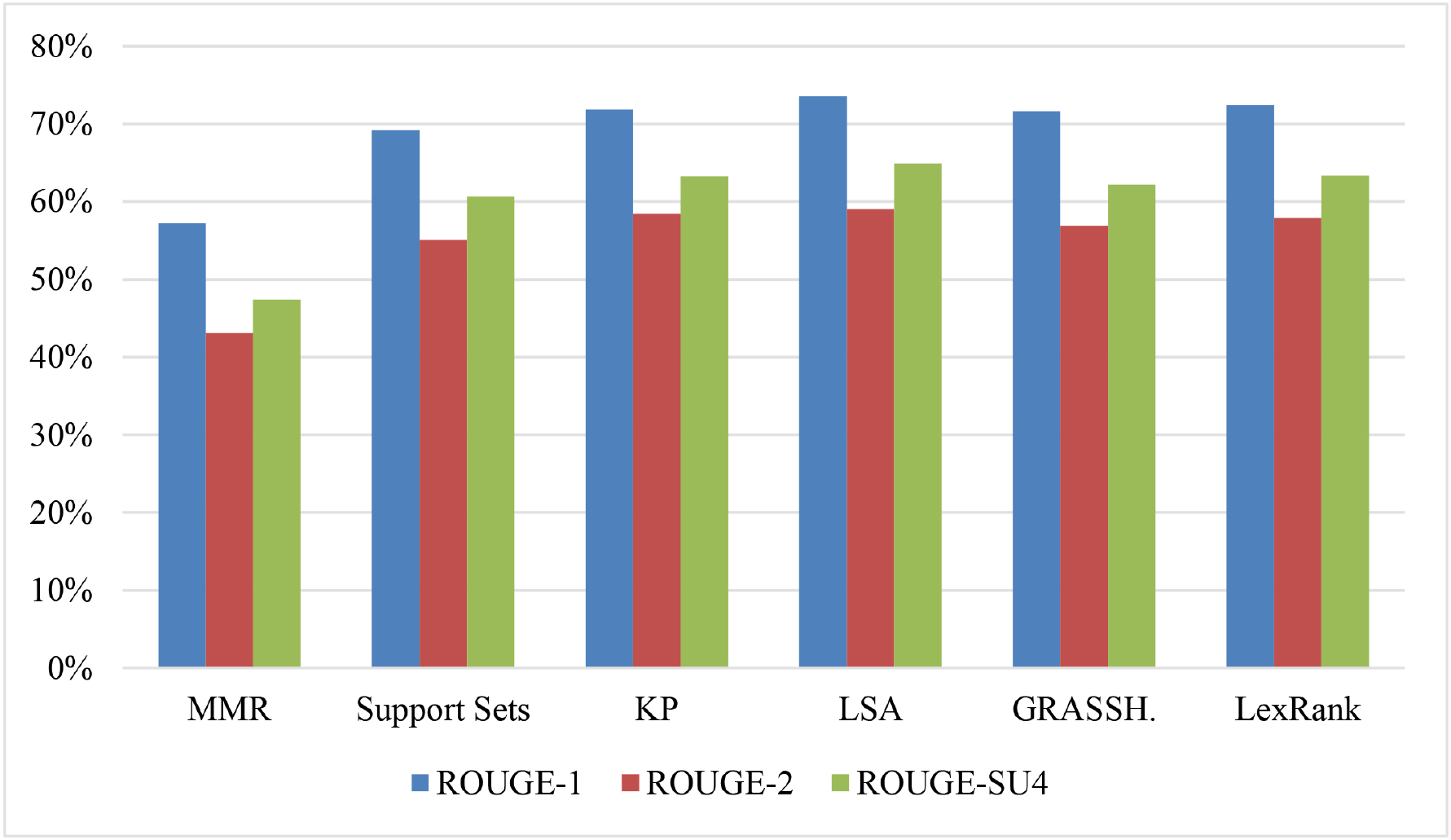}
	\end{center}
\caption{Relative performance for news articles. For MMR, $\lambda=0.50$; Support Sets used Manhattan distance and Support Set Cardinality = 2; KP-Centrality used 10 key phrases.}
	\label{fig:News_Alg}
\end{figure}

Table~\ref{table:News_ROUGE} presents the scores for each summarization algorithm. LSA achieved the best scores for R-1, R-2, and R-SU4. Figure~\ref{fig:News_Alg} shows the relative performance results.

\begin{table}[H]
\caption{ROUGE scores for generated summaries and original documents against manual references. For MMR, $\lambda=0.50$; Support Sets used Manhattan distance and Support Set Cardinality = 2; KP-Centrality used 10 key phrases.}
\label{table:News_ROUGE}
\centering
\begin{tabular}{lccc|c}
\hline
& R-1 & R-2 & R-SU4 & AVG \#Words \\\hline
MMR & 0.43 & 0.15 & 0.18 & 195\\
Support Sets & 0.52 & 0.19 & 0.23 & 254 \\
KP         & 0.54 & \textbf{0.20} & \textbf{0.24}  & 268
\\
LSA & \textbf{0.56} & \textbf{0.20} & \textbf{0.24} & 297\\
GRASSH. & 0.54 & 0.19 & 0.23 & 270 \\
LexRank & 0.55 & \textbf{0.20} & \textbf{0.24} & 277\\\hline
Original Docs & 0.75 & 0.34 & 0.38 & 608 \\ \hline

\end{tabular}
\end{table}

\subsection{Films}\label{sub:films-results}

Table~\ref{tab:resultsPlotSums} presents the scores for the film data combinations against plot summaries.
Overall, Support Sets, LSA, and LexRank, capture the most relevant sentences for plot summaries. It would be expected, for algorithms such as GRASSHOPPER and MMR, that maximize diversity, to perform well in this context, because plot summaries are relatively small and focus on the more important aspects of the film, ideally, without redundant content. However, our results show otherwise.  
For scripts, LSA and LexRank are the best  approaches in terms of R-1 and R-SU4.

\begin{table}[H]
\caption{ROUGE scores for generated summaries for subtitles, scripts, and scripts concatenated with subtitles, against plot summaries.  For MMR, $\lambda=0.50$; Support Sets used the cosine distance and threshold = $50\%$; KP-Centrality used 50 key phrases.}
\label{tab:resultsPlotSums}
\centering
\footnotesize
\begin{tabular}{p{.13\columnwidth}lccc|c}
 \hline                             &                    & R-1  & R-2     & R-SU4 & AVG \#Words \\\hline
\multirow{3}{*}{MMR}                & Subtitles          & 0.07 & 0.01    & 0.02 & 52 \\
                                    & Script             & 0.14 & 0.01    & 0.03  & 53

                                    \\
                                    & Script + Subtitles   & 0.12 & 0.01    & 0.03  & 71
\\\hline
\multirow{3}{*}{Support Sets}       & Subtitles          & 0.23 & \textbf{0.02}    & \textbf{0.06} & 150
 \\
                                    & Script             & 0.25 & 0.02    & 0.07 & 133
 \\
                                    & Script + Subtitles   & 0.29 & \textbf{0.03}    & \textbf{0.09} & 195
 \\\hline
 
 \multirow{3}{*}{KP}       & Subtitles          & 0.22 & 0.02 & \textbf{0.06} & 144 
 \\
                           & Script             & 0.24 & 0.02 & 0.07 & 123
 \\
                           & Script + Subtitles   & 0.28 & 0.02   & 0.08 & 184
 \\\hline
 
\multirow{3}{*}{LSA}                & Subtitles          & 0.22 & \textbf{0.02}    & \textbf{0.06} &  167
 \\
                                    & Script             & 0.28 & \textbf{0.03}    & 0.08 & 190
 \\
                                    & Script + Subtitles & 0.28 & \textbf{0.03}    & 0.08 & 219
 \\\hline
\multirow{3}{*}{GRASSH.}            & Subtitles          & 0.17 & 0.01    & 0.04 & 135
 \\
                                    & Script             & 0.21 & 0.02    & 0.06 & 121
 \\
                                    & Script + Subtitles & 0.22 & 0.02    & 0.06 &  118
\\\hline
\multirow{3}{*}{LexRank}            & Subtitles          & \textbf{0.24} & \textbf{0.02}    & \textbf{0.06} & 177
 \\
                                    & Script             & \textbf{0.29} & 0.02    & \textbf{0.09} & 168
 \\
                                    & Script + Subtitles & \textbf{0.30} & 0.02    & 0.08 &  217
\\\hline
\multirow{3}{*}{\vbox{Original\\ Docs}} & Subtitles          & 0.77 & 0.21    & 0.34  & 10460  \\
                                    & Script             & 0.74 & 0.23    & 0.36  & 14560  \\
                                    & Script + Subtitles & 0.83 & 0.31    & 0.43  & 24640  \\ \hline

\end{tabular}
\end{table}

\begin{table}[H]
\caption{ROUGE scores for generated summaries for subtitles, scripts, and scripts+subtitles, against plot synopses. For MMR, $\lambda=0.50$; Support Sets used the cosine distance and threshold = $50\%$; KP-Centrality used 50 key phrases.}
\label{tab:resultsSyn}
\centering
\footnotesize
\begin{tabular}{p{.10\columnwidth}lccc|c}
\hline
                                    &                    & R-1  & R-2  & R-SU4   &  AVG \#Words  \\\hline
\multirow{3}{*}{MMR}                & Subtitles          & 0.08 & 0.01 & 0.02  &  435
  \\
                                    & Script             & 0.16 & 0.03 & 0.06  &  745
\\
                                    & Script + Subtitles & 0.11 & 0.01 & 0.03  &  498
  \\\hline
\multirow{3}{*}{\vbox{Support\\ Sets}}       & Subtitles & 0.25 & 0.04 & 0.08  &  1033
  \\
                                    & Script             & 0.37 & 0.07 & 0.15  &  1536
\\
                                    & Script + Subtitles & 0.42 & 0.08 & 0.16  &  1736
   \\\hline
   
    \multirow{3}{*}{KP}       & Subtitles          & 0.24 & 0.04 & 0.08 & 952
    \\
    & Script                                       & 0.36 & 0.07 & 0.14 & 1419
    \\
    & Script + Subtitles                           &  0.40 & 0.08 & 0.16 & 1580
    \\\hline
\multirow{3}{*}{LSA}                & Subtitles          & 0.31 & \textbf{0.06} & 0.11  & 1303
   \\
                                    & Script             & 0.42 & 0.09 & 0.17   &  1934
 \\
                                    & Script + Subtitles & 0.45 & \textbf{0.10} & 0.18   &  2065
 \\\hline
\multirow{3}{*}{GRASSH.}            & Subtitles          & \textbf{0.34} & \textbf{0.06} & \textbf{0.12}  &  1553
  \\
                                    & Script             & 0.44 & 0.09 & \textbf{0.18}   & 1946
  \\
                                    & Script + Subtitles & 0.47 & \textbf{0.10} & \textbf{0.19}   &  1768
 \\\hline
\multirow{3}{*}{LexRank}            & Subtitles          & \textbf{0.34} & \textbf{0.06} & \textbf{0.12}   & 1585
 \\
                                    & Script             & \textbf{0.45} & \textbf{0.10} & \textbf{0.18}   &  1975
 \\
                                    & Script + Subtitles & \textbf{0.48} & \textbf{0.10} & \textbf{0.19}   &   2222
\\\hline
\multirow{3}{*}{\vbox{Original\\ Docs}} & Subtitles          & 0.70 & 0.18 & 0.30  & 10460    \\
                                    & Script             & 0.73 & 0.24 & 0.37   & 14560    \\
                                    & Script + Subtitles & 0.83 & 0.32 & 0.44   & 24640   \\
\hline

\end{tabular}
\end{table}

Table~\ref{tab:resultsSyn} presents the scores for the film data combinations against plot synopses. The size of synopses is very different from that of plot summaries. Although synopses also focus on the major events of the story, their larger size allows for a more refined description of film events. Additionally, because summaries are created with the same number of sentences of the corresponding synopsis, higher scores are expected. From all algorithms, LexRank clearly stands out with the highest scores for all metrics (except for R-SU4, for scripts).

The script+subtitles combination was used in order to determine whether the inclusion of redundant content would improve the scores, over the separate use of scripts or subtitles. However, in all cases (Figure~\ref{fig:GraphConcl1}), script+subtitles leads to worse scores, when compared to scripts alone.
The same behavior is observed when using subtitles except for Support Sets-based methods (Support Sets and KP-Centrality).
%
For plot synopses, the best scores are achieved by LexRank and GRASSHOPPER, while, for plot summaries, the best scores are achieved by LexRank and LSA. By inspection of the summaries produced by each algorithm, we observed that MMR chooses sentences with fewer words in comparison with all other algorithms (normally, leading to lower scores). Overall, the algorithms behave similarly for both subtitles and scripts.

\subsection{Documentaries}\label{sub:documentaries-results}

\begin{figure}[H]
\begin{center}
\includegraphics[keepaspectratio=true,width=\columnwidth]{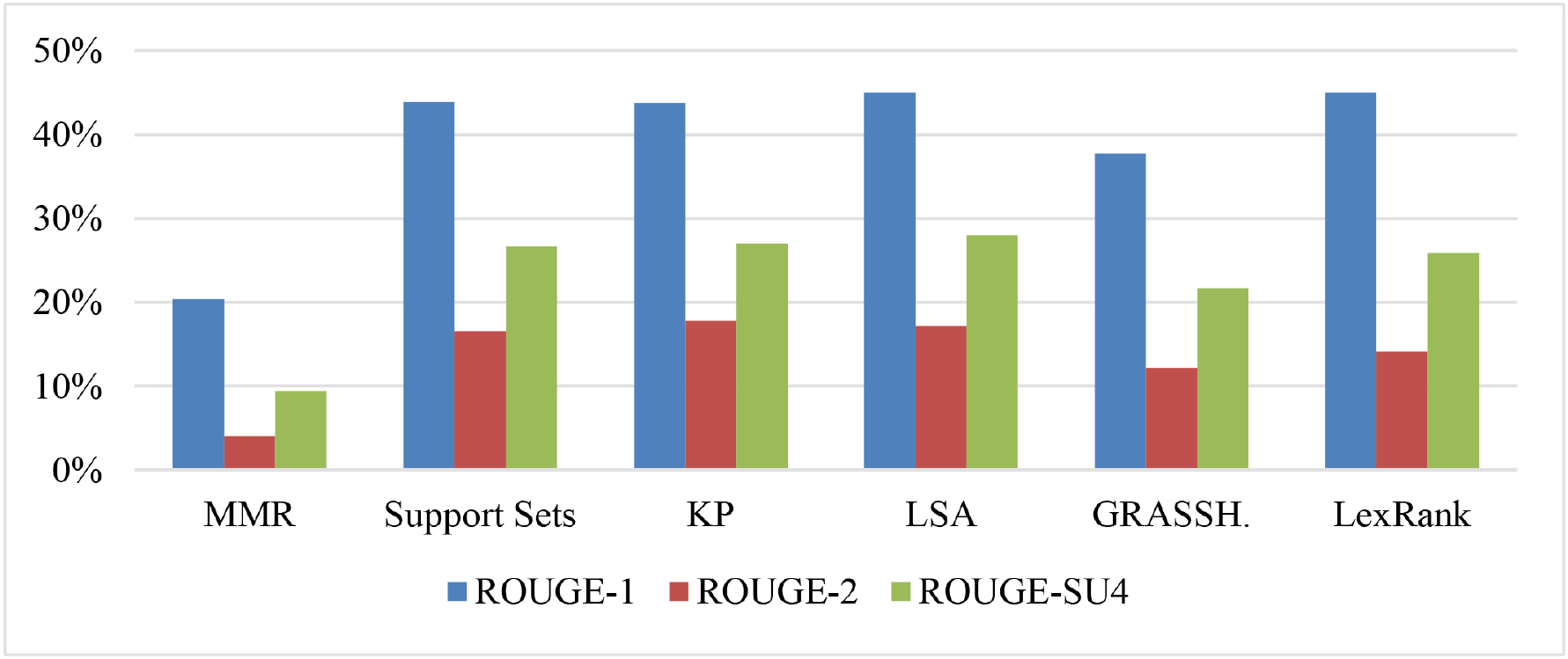}
\end{center}
\caption{Relative performance for documentaries against plot summaries. For MMR, $\lambda=0.75$; Support Sets used cosine distance and threshold=$50\%$; KP-Centrality used 50 key phrases.}
\label{fig:Doc_Alg}
\end{figure}

From all algorithms (Table~\ref{table:Doc_ROUGE}), LSA achieved the best results for R-1 and R-SU4, along with LexRank for R-1. KP-Centrality achieved the best results for R-2. It is important to notice that LSA also produces the summaries with the highest word count (favoring recall). Figure~\ref{fig:Doc_Alg} shows the relative performance  results: LSA outperformed all other algorithms for R-1 and R-SU4, and KP-Centrality was the best for R-2; Support Sets and KP-Centrality performed closely to LSA for R-SU4; the best MMR results were consistently worse across all metrics (MMR summaries have the lowest word count). 

\begin{table}[H]
\caption{ROUGE scores for generated summaries and   original subtitles against human-made plot summaries. For MMR, $\lambda=0.75$; Support Sets used the cosine distance and threshold = $50\%$; KP-Centrality used 50 key phrases.}
\label{table:Doc_ROUGE}
\centering
\begin{tabular}{lccc|c}
\hline
                     & R-1 & R-2  & R-SU4   &  AVG \#Words\\\hline
MMR                  & 0.17 & 0.01 & 0.04  & 78
\\
Support Sets         & 0.37 & 0.06 & 0.12  & 158
\\
KP         & 0.37 & \textbf{0.07}  & 0.12 & 149 
\\
LSA                  & \textbf{0.38} & 0.06 & \textbf{0.13} &  199
\\
GRASSH.          & 0.31 & 0.04 & 0.10  & 150
\\
LexRank              & \textbf{0.38} & 0.05 & 0.12 & 183
 \\ \hline
Original Docs   & 0.83 & 0.37 & 0.46  & 5864\\
\hline

\end{tabular}
\end{table}

\subsection{Discussion}\label{sec:discussion}

\begin{figure}[H]
	\begin{center}
		\includegraphics[keepaspectratio=true,width=\columnwidth]{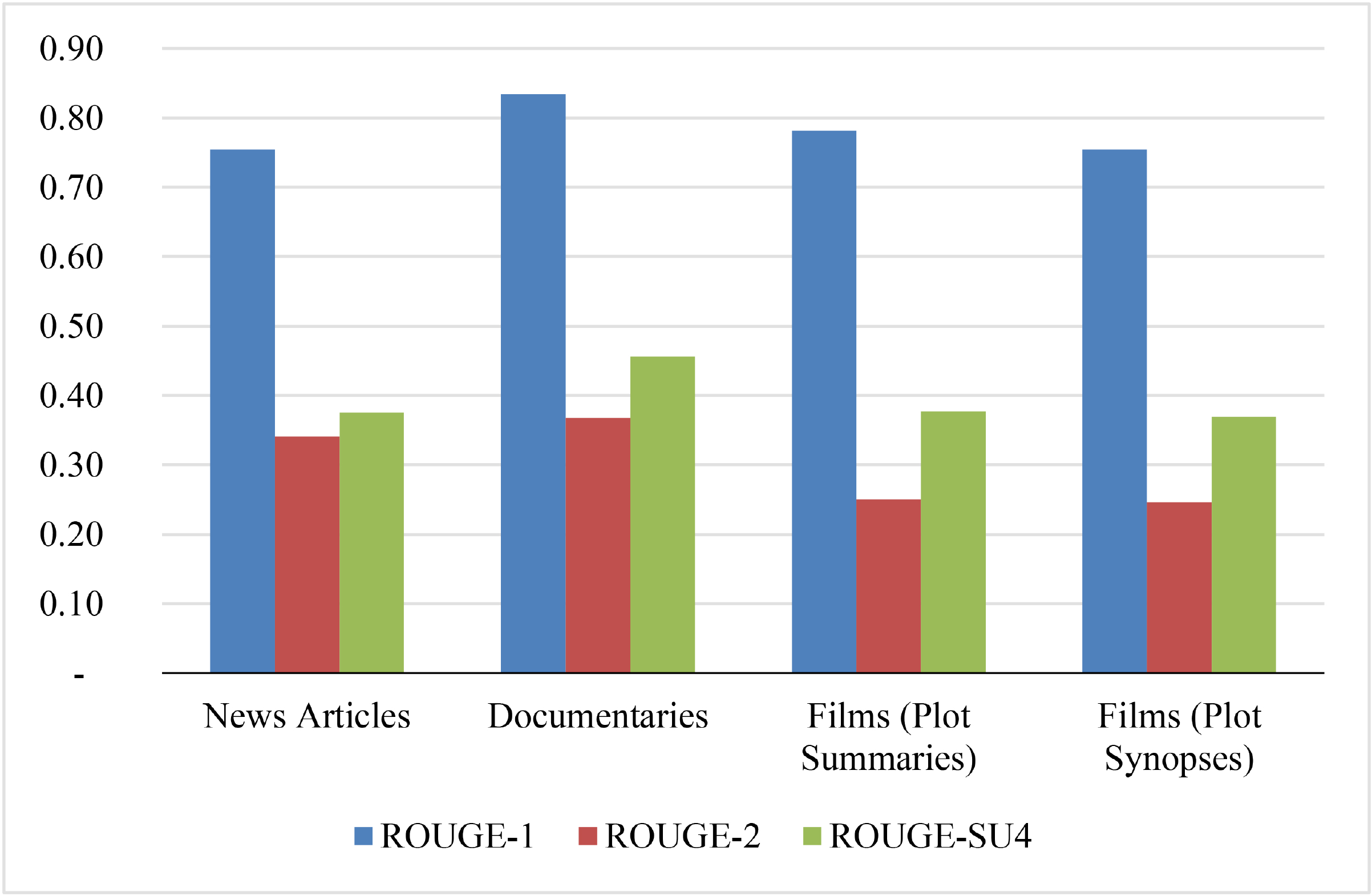}
	\end{center}
	\caption{ROUGE scores for news articles, films, and documentaries against manual references, plot summaries and synopses, and plot summaries, respectively.}
	\label{fig:Originais}
\end{figure}

News articles intend to answer basic questions about a particular event: who, what, when, where, why, and often, how. Their structure is sometimes referred to as  ``inverted pyramid", where the most essential information comes first. Typically, the first sentences provide a good overview of the entire article and are more likely to be chosen when composing the final summary. Although documentaries follow a narrative structure similar to films, they can be seen as more closely related to news than films, especially regarding their intrinsic informative nature. In spite of their different natures, however, summaries created by humans produce similar scores for all of them. It is possible to observe this behavior in Figure~\ref{fig:Originais}.
Note that documentaries achieve higher scores than news articles or films, when using the original subtitles documents against the corresponding manual plot summaries. 

\begin{figure*}
	\begin{center}
		\includegraphics[keepaspectratio=true,width=530px]{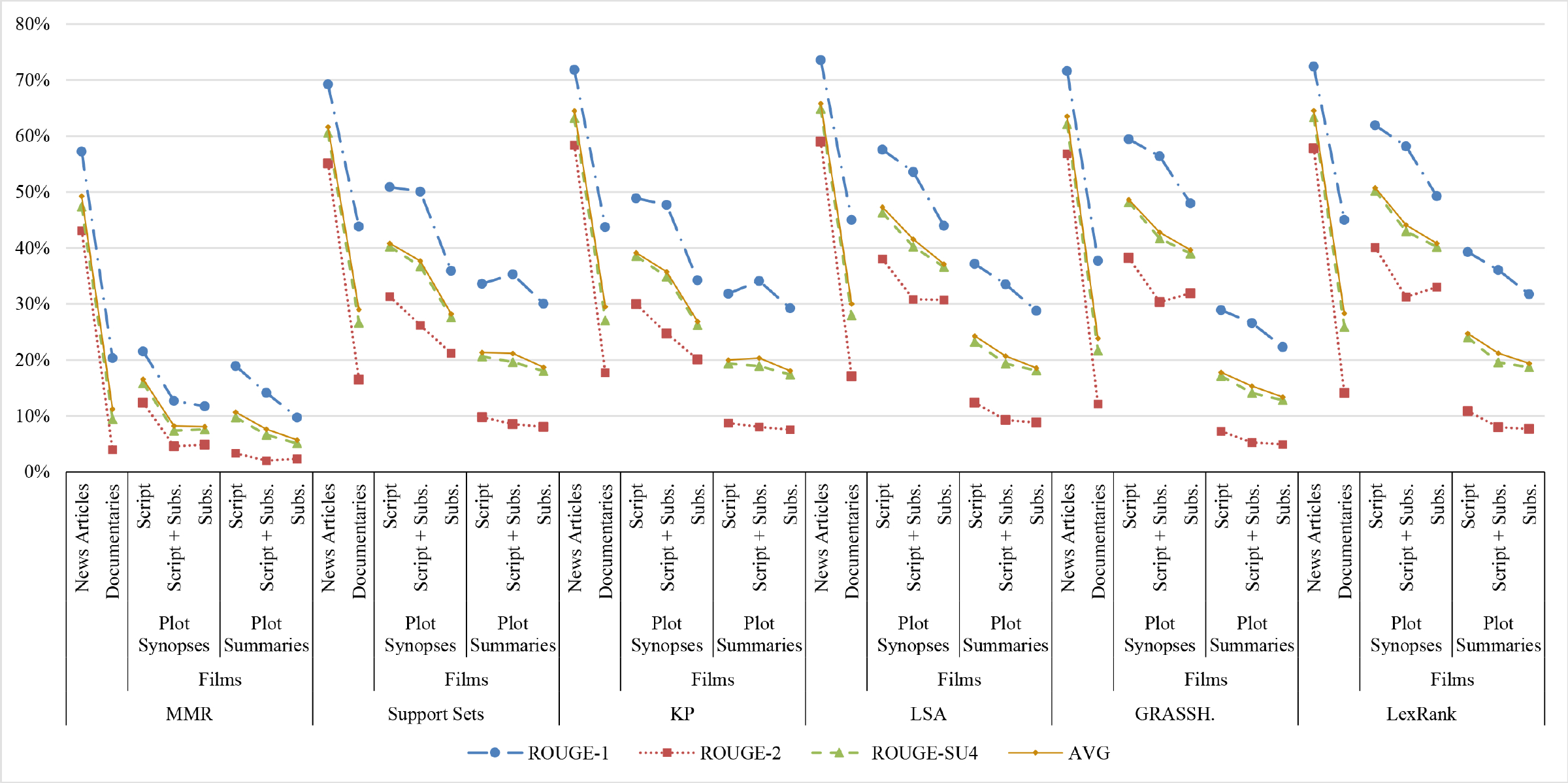}
	\end{center}
	\caption{Relative performance for all datasets. For films the relative performance was measured against plot synopses and plot summaries: MMR used $\lambda = 0.50$; and  Support Sets used the cosine distance and threshold = $50\%$; KP-Centrality used 50 key phrases.}
	\label{fig:GraphConcl1}
\end{figure*}

Figure~\ref{fig:GraphConcl1} presents an overview of the performance of each summarization algorithm across all domains. 
The results concerning news articles were the best out of all three datasets for all experiments. However, summaries for this dataset preserve, approximately, 31\% of the original articles, in terms of sentences, which is significantly higher than for films and documentaries (which preserve less than $1\%$), necessarily leading to higher scores. Nonetheless, we can observe the differences in behavior between these domains. Notably, documentaries achieve the best results for plot summaries, in comparison with films, using scripts, subtitles, or the combination of both.
The relative scores on the films dataset are influenced by two major aspects: the short sentences found in the films dialogs; and, since the generated summaries are extracts from subtitles and scripts, they are not able to represent the film as a whole, in contrast with what happens with plot summaries or synopses.
Additionally, the experiments conducted for script+subtitles for films, in general, do not improve scores above those of scripts alone, except for Support Sets for R-1. 
Overall, LSA performed consistently better for news articles and documentaries. Similar relatively good behavior had already been observed for meeting recordings, where the best summarizer was also LSA~\citep{murray:renals:carletta:2005}. One possible reason for these results is that LSA tries to capture the relation between words in sentences. By inferring contextual usage of text based on these relations, high scores, apart from R-1, are produced for R-2 and R-SU4. For films, LexRank was the best performing algorithm for subtitles, scripts and the combination of both, using plot synopses, followed by LSA and Support Sets for plot summaries. MMR has the lowest scores for all metrics and all datasets. We observed that sentences closer to the centroid typically contain very few words, thus leading to shorter summaries and the corresponding low scores.

Interestingly, by observing the average of R-1, R-2, and R-SU4, it is possible to notice that it follows very closely the values of R-SU4. These results suggest that R-SU4 adequately reflects the scores of both R-1 and R-2, capturing the concepts derived from both unigrams and bigrams.

Overall, considering plot summaries, documentaries achieved higher results in comparison with films. However, in general, the highest score for these two domains is achieved using films scripts against plot synopses. Note that synopses have a significant difference in terms of sentences in comparison with plot summaries. The average synopsis has 120 sentences, while plot summaries have, on average, 5 sentences for films, and 4 for documentaries. This gives synopses a clear advantage in terms of ROUGE (recall-based) scores, due to the high count of words.

\section{Conclusions and Future Work}\label{sec:conclusions}

We analyzed the impact of the six summarization algorithms on three datasets. The newspaper articles dataset was used as a reference. The other two datasets, consisting of films and documentaries, were evaluated against plot summaries, for films and documentaries, and synopses, for films. Despite the different nature of these domains, the abstractive summaries created by humans, used for evaluation, share similar scores across metrics.

The best performing algorithms are LSA, for news and documentaries, and LexRank for films. Moreover, we conducted experiments combining scripts and subtitles for films, in order to assess the performance of generic algorithms by inclusion of redundant content. Our results suggest that 
this combination is unfavorable. Additionally, it is possible to observe that all algorithms behave similarly for both subtitles and scripts. As previously mentioned, the average of the scores follows closely the values of R-SU4, suggesting  that R-SU4 is able to capture concepts derived from both unigrams and bigrams.

We plan to use subtitles as a starting point to perform video summaries of films and documentaries. For films, the results from our experiments using plot summaries show that the summarization of scripts only marginally improved performance, in comparison with subtitles. This suggests that subtitles are a viable approach for text-driven film and documentary summarization. This positive aspect is compounded by their being broadly available, as opposed to scripts. 

\section{Acknowledgements}

This work was supported by national funds through Fundação para a Ciência e a Tecnologia (FCT) with reference \linebreak UID/CEC/50021/2013.


\bibliographystyle{model2-names}

\end{document}